\documentclass[journal,twoside,web]{ieeecolor}
\usepackage{tmi}
\usepackage{cite}
\usepackage{amsmath,amssymb,amsfonts}

\usepackage{algorithmic}
\usepackage{graphicx}
\usepackage{textcomp}
\usepackage{makecell, multirow}
\usepackage{algorithm2e}
\usepackage[table]{xcolor}
\usepackage{threeparttable}
\usepackage[nodisplayskipstretch]{setspace}

\definecolor{zolty}{RGB}{255,236,184}
\definecolor{rozowy}{RGB}{251,232,228}
\definecolor{niebieski}{RGB}{215,238,247}
\definecolor{fioletowy}{RGB}{254,154,254}
\definecolor{zielony}{RGB}{217,255,219}
\definecolor{szary}{RGB}{229,229,229}

\def\BibTeX{{\rm B\kern-.05em{\sc i\kern-.025em b}\kern-.08em
    T\kern-.1667em\lower.7ex\hbox{E}\kern-.125emX}}
\begin{document}

\title{Implantable Adaptive Cells: A Novel Enhancement for Pre-Trained U-Nets in Medical Image Segmentation}
\author{Emil Benedykciuk, Marcin Denkowski, and Grzegorz M. W\'{o}jcik, 
\thanks{Emil Benedykciuk is with the Institute of Computer Science, Maria Curie-Sk\l{}odowska University, Lublin, 20-033, Poland (e-mail:emil.benedykciuk@umcs.pl). }
\thanks{Marcin Denkowski is with the Institute of Computer Science, Maria Curie-Sk\l{}odowska University, Lublin, 20-033, Poland (e-mail:marcin.denkowski@umcs.pl). }
\thanks{Grzegorz M. W\'{o}jcik is with the Institute of Computer Science, Maria Curie-Sk\l{}odowska University, Lublin, 20-033, Poland (e-mail:gmwojcik@umcs.edu.pl). }
}

\maketitle

\begin{center}
	\textbf{Preprint Notice:} This manuscript has been submitted to IEEE Transactions on Medical Imaging for review. The final published version may differ.
\end{center}

\begin{abstract}
    This paper introduces a novel approach to enhance the performance of pre-trained neural networks in medical image segmentation using gradient-based Neural Architecture Search (NAS) methods.
    We present the concept of Implantable Adaptive Cell (IAC), small modules identified through Partially-Connected DARTS based approach, designed to be injected into the skip connections of an existing and already trained U-shaped model.
	Unlike traditional NAS methods, our approach refines existing architectures without full retraining.
	Experiments on four medical datasets with MRI and CT images show consistent accuracy improvements on various U-Net configurations, with segmentation accuracy gain by approximately 5 percentage points across all validation datasets, with improvements reaching up to 11\%pt in the best-performing cases.
    The findings of this study not only offer a cost-effective alternative to the complete overhaul of complex models for performance upgrades but also indicate the potential applicability of our method to other architectures and problem domains.
	The project and code are available at https://gitlab.com/emil-benedykciuk/u-net-darts-tensorflow.
\end{abstract}


\section{Introduction}\label{sec:introduction}

\IEEEPARstart{D}{esigning} neural network architectures remains a challenge in deep learning, with performance influenced by depth, operations, and normalization. Neural Architecture Search (NAS) automates this process, optimizing structures efficiently.
NAS, originally used in classification \cite{bib:nas_review}, has been adapted for segmentation tasks \cite{bib:AutoDeepLab, bib:FasterSeg, bib:DCNAS}, including medical imaging \cite{bib:AutoCNN, bib:NAS-Unet, bib:RONASMIS, bib:V-NAS, bib:DiNTS, bib:HyperSegNAS}. 
Many of these studies employ gradient-based NAS, particularly Differentiable Architecture Search (DARTS) \cite{bib:DARTS} and its variants \cite{bib:DARTS+, bib:PC-DARTS, bib:DropNAS, bib:SingleDARTS}, due to the computational demands of high-resolution data. 
However, existing approaches typically yield entirely new architectures that must be trained from scratch with large datasets \cite{bib:NAS-Unet, bib:DiNTS}.

\begin{figure*}[ht]
	\begin{center}
		\includegraphics[width=0.75\textwidth]{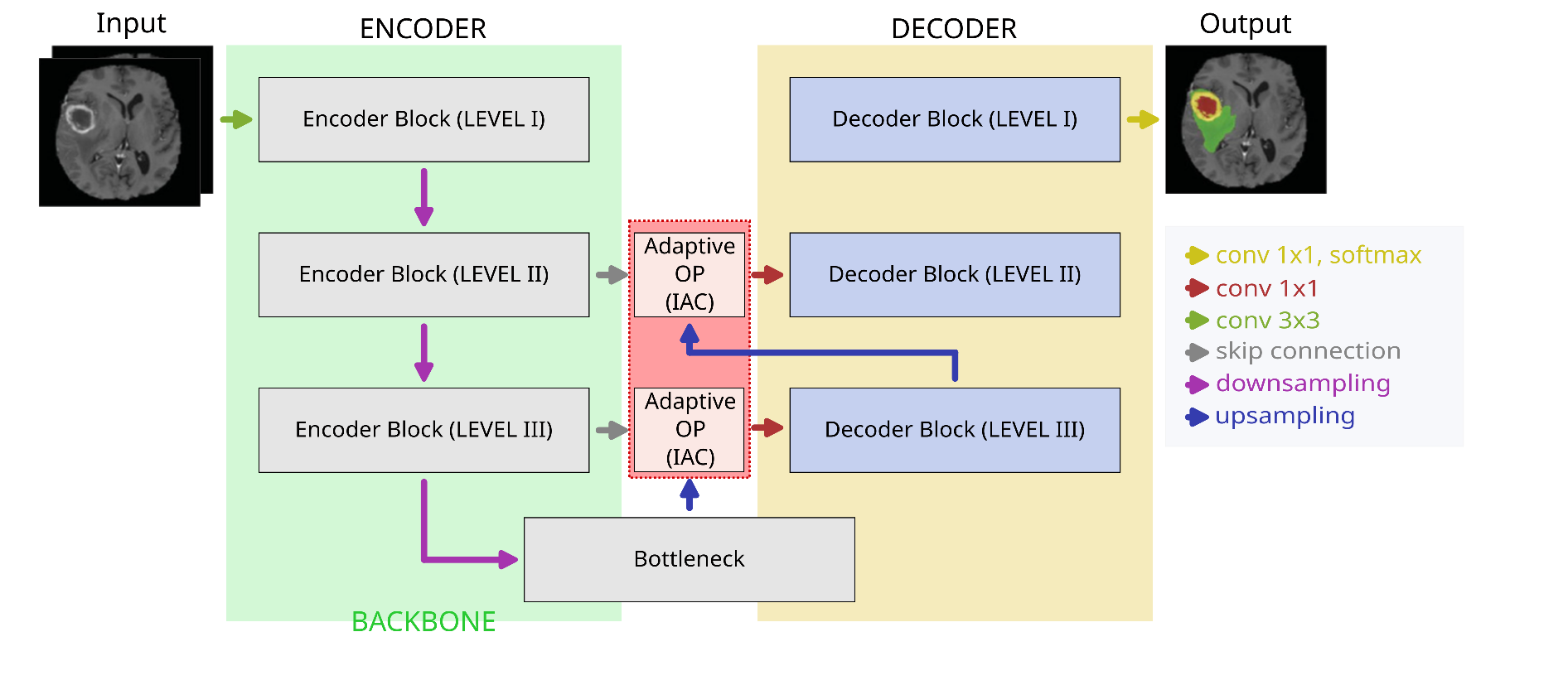} 
	\end{center}
	\caption{Diagram of the U-Net with Implantable Adaptive Cells (IAC) integration. It is worth noting that the BACKBONE is intentionally varied. \textbf{Adaptive OP} varies by stage: concat (Stage I), continuous IAC (Stage II), discrete IAC (Stage III). Detailed diagram of IAC is shown in Fig.~\ref{fig:cell}.}
	\label{fig:unet_general}
\end{figure*}

A gap remains in leveraging gradient-based NAS for improving pre-trained networks. 
No prior work has examined integrating NAS-generated Cells into U-Net skip connections, offering a way to refine models without full retraining—a crucial aspect in medical imaging, where data and computing power are often limited.
To address this gap, we propose refining the skip connections of pre-trained U-Nets, critical for transferring features between encoder and decoder, by identifying small blocks called \emph{Implantable Adaptive Cells (IAC)} (see Fig. \ref{fig:unet_general}). 
Unlike attention mechanisms, IACs use only convolution and pooling operations, ensuring efficiency. 
We adopt a differentiable NAS approach inspired by PC-DARTS \cite{bib:PC-DARTS} to discover optimal IAC architectures. 
While they share the same cell structure at each resolution level of the U-Net, the learned weights differ, allowing multi-scale feature extraction.
This approach offers several advantages:
\begin{enumerate}
	\item \textbf{Improved Feature Fusion}: By identifying optimal cells for skip connections, we enhance the feature fusion process, leading to superior segmentation performance.
	\item \textbf{Resource Efficiency}: Integrating IACs with an existing model is computationally more efficient than searching or training a new model from scratch, which is particularly advantageous for resource-constrained institutions.
	\item \textbf{Task-Specific Adaptation}: Different medical imaging tasks may benefit from distinct skip-connection operations. NAS enables the automatic identification of the most appropriate operations tailored to specific tasks.
	\item \textbf{Leverage of Pre-Trained Models}: Utilizing pre-trained models reduces training time and capitalizes on existing learned representations, which is especially beneficial in scenarios with limited data.
\end{enumerate}

Integrating NAS with pre-trained models can introduce challenges, such as potential conflicts with existing weights and disruption of learned representations. Our aim is not necessarily to surpass state-of-the-art methods in every segmentation task, but to offer a general method for enhancing pre-trained U-Nets by embedding small, adaptable modules into their skip connections.
The main contributions of our approach include:
\begin{enumerate}
	\item \textbf{Innovative Application of NAS}: We introduce a novel use of differentiable NAS methods to optimize cells within skip connections of pre-designed and pre-trained U-Net architectures. This alternative approach significantly reduces the time to find a solution achieving acceptable results for the selected task.
	\item \textbf{Development of Implantable Adaptive Cell (IAC)}: We propose a method for generating IACs, compatible with any encoder architecture in U-Nets, enhancing their flexibility.
	\item \textbf{Enhanced already trained U-Net Performance}: By integrating IACs into the skip connections of pre-trained U-Nets, we improve their efficiency and generalization capabilities.
	\item \textbf{Feature Processing with IAC}: IAC processes both low- and high-level features, enhancing the network's overall learning capacity.
	\item \textbf{Experimental Validation}: We conduct comprehensive experiments on standard medical imaging datasets (ACDC \cite{bib:acdc}, BraTS2023 \cite{bib:BraTS-2015,bib:BraTS-Dodatkowy1,bib:BraTS-2023}, KiTS2023 \cite{bib:kits21}, AMOS2022 \cite{bib:amos}), showing performance improvements over baseline U-Net models.
	\item \textbf{Challenges and Solutions in NAS Integration}: We discuss the challenges of integrating NAS with pre-trained models and provide insights into overcoming these obstacles.
\end{enumerate}

To facilitate further research and application, we have made the source code of our implementation publicly available at: \\ {https://gitlab.com/emil-benedykciuk/u-net-darts-tensorflow}.

\section{Related Work}

In recent years, deep neural networks have become a pivotal tool in automating the segmentation of medical images. Despite substantial progress, accurately distinguishing pathological changes from healthy tissues in MRI images remains challenging and often requires specialized architectures.

A major breakthrough in biomedical image analysis was the U-Net \cite{bib:unet_base}, organized in an encoder-decoder structure. The encoder extracts features at increasingly higher levels of abstraction, while the decoder reconstructs these features back to higher resolution, producing dense prediction maps. This design inspired numerous modifications: Attention U-Net \cite{bib:AttentionUNet} and RAUnet \cite{bib:RAUnet} introduce attention mechanisms, UNet++ \cite{bib:UNet++} leverages dense skip connections, and nnU-Net \cite{bib:nnUNet,bib:nnUNet2024} automatically determines optimal configurations for various tasks, winning challenges such as the Medical Segmentation Decathlon \cite{bib:MSD_challenge2019}.
Skip connections play a crucial role in U-Net, transferring high-resolution feature maps from the encoder to the decoder and preserving spatial information essential for segmenting small structures. However, these connections can also introduce over- or under-segmentation errors. To address this, techniques such as attention gates, additional convolutional layers, or dense blocks refine feature transfer by discarding irrelevant regions and emphasizing task-relevant features \cite{bib:AttentionUNet,bib:RAUnet,bib:UNet++}.

\textbf{Neural architecture search (NAS)} automates network design, evolving from reinforcement learning \cite{bib:RF1} and evolutionary methods \cite{bib:Evolution1} to more efficient DNAS like DARTS  \cite{bib:DARTS}.
However, challenges such as instability and performance collapse led to improvements like PDARTS \cite{bib:PDARTS} and PC-DARTS \cite{bib:PC-DARTS}, SingleDARTS \cite{bib:SingleDARTS} or $\ell$-DARTS \cite{bib:lDarts}.
The last one \cite{bib:lDarts} proposed a compensation module and margin-valued regularization, reducing the bias toward parameter-free operations.
A recent comprehensive survey on NAS in medical imaging \cite{bib:nas_taxonomy} categorizes various NAS approaches, highlighting key trends and challenges in adapting NAS for biomedical tasks.
The authors in  \cite{bib:nas_biomedical} suggest that architecture search methods can provide significant improvements across different imaging modalities for classification task. 
Advancing, NAS has been applied to both natural image segmentation \cite{bib:AutoDeepLab, bib:FasterSeg, bib:DCNAS} and medical imaging \cite{bib:AutoCNN, bib:NAS-Unet, bib:RONASMIS}, often refining U-Net architectures \cite{bib:V-NAS} or incorporating broader search spaces \cite{bib:AutoDeepLab, bib:FasterSeg}. DiNTS \cite{bib:DiNTS} employed NAS for 3D segmentation and performed strongly in the 2021 Medical Segmentation Decathlon \cite{bib:MSD_challenge2022}, while HyperSegNAS \cite{bib:HyperSegNAS} introduced a Meta-Assistant Network for adaptive channel weighting. 

Current research largely focuses on fully automated NAS-generated architectures or manually designed U-Net enhancements (e.g., attention blocks in skip connections). 
Both approaches typically require training from scratch, which can be time-consuming and resource-intensive. 
Systems like nnU-Net \cite{bib:nnUNet,bib:nnUNet2024} or Auto3DSeg \cite{bib:DiNTS} partially automate architecture selection or use model ensembles, but they do not seamlessly integrate new elements into an already trained network.

To date, no attempt has been made to use NAS for generating additional modules (e.g., within skip connections) that adapt to a previously designed, pre-trained architecture. Such flexibility could significantly shorten the time and cost of improving model accuracy or adapting the network to new data, a critical factor in clinical settings. In this work, we propose leveraging a pre-trained U-shaped network and creating small adaptive modules that can be inserted into the existing architecture. This aims to reduce the computational and temporal overhead compared to approaches that generate and train a full network from scratch using NAS.

\section{Method}\label{sec:method}

Research suggests that applying differentiable architecture search to skip connections can help discover better network structures. 
This can further improve the network's ability to learn complex features while still preserving spatial information.
It is important to note that traditional NAS methods mainly focus on tuning hyperparameters and defining the supernetwork, which researchers design by specifying the number and arrangement of cells. 
In our approach, an additional challenge is that the search process must adapt to an already pre-trained and pre-designed network. 
This distinction directly impacts the optimization problem and the loss function landscape.


\subsection{Base network topology}
In our research, we use 2D U-Net architecture \cite{bib:unet_base}, referred to as base U-Net.
Our decisions are guided by experiences with nnU-Net presented in \cite{bib:nnUNet,bib:nnUNet2024}, where researchers take away superfluous elements of many proposed network designs.
To evaluate the generated IAC, we train various U-shaped architectures with modifications to the encoder.
Specifically, we replace the base architecture with different backbones, such as EfficientNetV2 \cite{bib:effv2}, ConvNeXt \cite{bib:ConvNext}, MobileNetV3 \cite{bib:mobilenetV3} and others, while keeping the U-shaped structure intact.
The ConvNeXt \cite{bib:ConvNext} architecture was chosen due to the significant successes of MedNeXt \cite{bib:MedNeXt}, which was also shown by the authors of nnU-Net \cite{bib:nnUNet2024}, where this solution achieved the best results for 3D segmentation across all datasets (except KiTS \cite{bib:kits21}).
This allows us to assess the performance and compare the efficacy of our cell in different U-Net frameworks.
This analysis helps us understand the influence of the cell architecture on the overall performance of the U-shape models in our specific medical image segmentation task.


\subsection{Implantable Adaptive Cell architecture}\label{sec:cell_arch}

\begin{figure}[ht]
	\begin{center}
		\includegraphics[width=0.95\columnwidth]{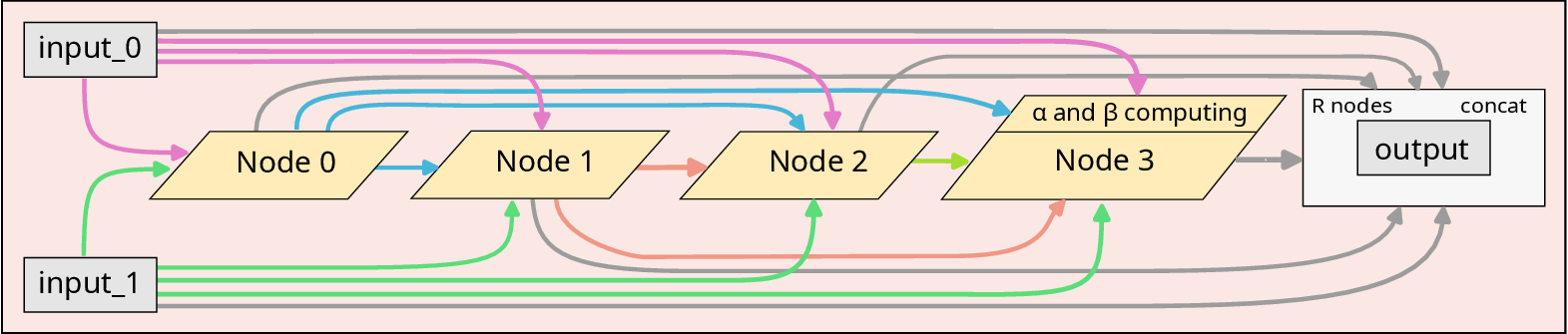} 
	\end{center}
	\caption{IAC diagram: \textbf{input\textsubscript1} are coarse features procured via skip connections from the encoder, and \textbf{input\textsubscript0} are upsampled decoder features. Search explores node connections and operations. A visual representation of the node block (yellow) is shown in Fig.~\ref{fig:pc_darts}.}
	\label{fig:cell}
\end{figure}

We define our Implantable Adaptive Cell (IAC) as a compact directed acyclic graph (DAG) consisting of $\mathcal{N}$ nodes, where each node represents a hidden feature representation.
The transformation of feature tensors between a pair of nodes is performed by edges, which correspond to the selected operations to be executed.
Each edge encompasses $\mathcal{M}$ candidate operations, offering a diverse array of potential transformations (see~Fig.~\ref{fig:cell}).
Notably, each edge and its associated operation have independent weights, called architecture parameters, discussed further in Section~\ref{sec:relaxation}.
The cell functions as a fully convolutional module, where the input for each node results from a weighted sum of operations originating from all previous nodes. It's important to note that we are still discussing the continuous architecture representation of the cell at this stage.

Our Implantable Adaptive Cell (IAC) is strategically placed within the U-Net architecture, with a single architecture shared across all levels, while individual cell weights are trained separately. 
These weights correspond to trainable operations along graph edges and input/output transformations. 
The IAC is applied across all skip connections, adapting at different levels with distinct weights and operations.
Unlike conventional NAS cell searches, our IAC integrates features from both the encoder and decoder of a frozen U-Net, posing additional challenges for feature fusion and optimization.
 Positioned within skip connections, the cell receives coarse encoder features and upsampled decoder outputs. 
To manage different feature dimensions, a $1\times1$ convolution is applied before processing, and another $1\times1$ convolution aligns the final output.
The continuous representation of the IAC is used to determine the optimal cell structure, which is later discretized for final implementation.


\subsubsection{Search space}
The search space consists of fundamental operations that preserve the spatial resolution of input and output feature maps. 
The search space is limited to operations preserving spatial resolution, ensuring efficient NAS exploration in U-shaped architectures.

For this study, we adopt the operations from DARTS \cite{bib:DARTS} and PC-DARTS \cite{bib:PC-DARTS}, as our focus is on evaluating DNAS methods for pre-trained architectures. The simplicity of these operations is sufficient for this purpose.

Each cell comprises four intermediate nodes connected by 14 edges, with each edge offering a choice of 8 operations: zero (no connection), identity (skip connection), $3\times3$ max pooling, $3\times3$ average pooling, $3\times3$ and $5\times5$ separable convolutions, and $3\times3$ and $5\times5$ dilated separable convolutions.


\subsubsection{Continuous relaxation and optimization}\label{sec:relaxation}

PC-DARTS employs continuous relaxation to facilitate operation selection between nodes in the Adaptive Cell. 
Following the DARTS and PC-DARTS approaches, a mixed operation (softmax-based combination) enables gradient-based optimization. 
Given an operation set $\mathcal{O}$, each edge $(i, j)$ in a cell is represented as a weighted sum of possible operations:
\begin{align}
\varphi_o = \frac{\exp (\alpha^o_{i,j})}{\sum_{o' \in \mathcal{O}} \exp(\alpha^{o'}{i,j})} \label{eq:1} \\
f_{i,j}(x_i) = \sum_{o\in \mathcal{O}} \varphi_o \cdot o(x_i) \label{eq:2}
\end{align}
where $x_i$ is the output of node $i$, and $\alpha^o_{i,j}$ represents trainable architecture parameters.

\begin{figure}[ht]
	\begin{center}
		\includegraphics[width=0.98\columnwidth]{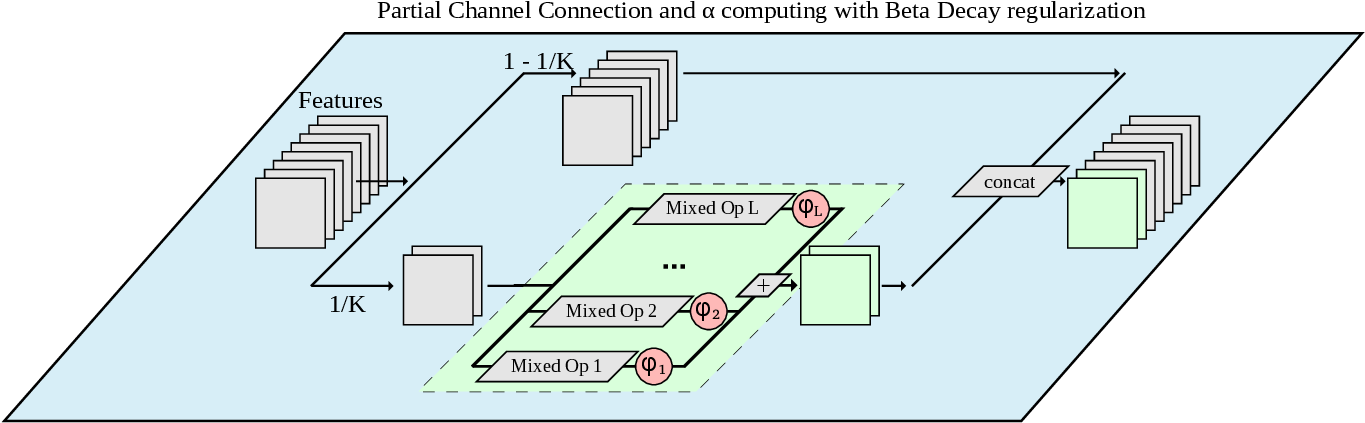} 
	\end{center}
	\caption{A visual representation of a cell architecture search by continuously relaxing the search space by placing a mixture of candidate operations on each edge.}
	\label{fig:alphas_beta}
\end{figure}

To improve memory efficiency, PC-DARTS introduces partial channel connection, where only a fraction ($1/\mathcal{K}$) of channels participates in mixed computation. A channel mask $\mathcal{M}_{i,j}$ selects active channels, while inactive ones bypass operations:
\begin{equation}
f^{PC}_{i,j}(x_i; \mathcal{M}_{i,j}) = \sum_{o\in \mathcal{O}}{(\varphi_o \cdot o(\mathcal{M}_{i,j}x_i))} + (1-\mathcal{M}_{i,j})x_i
\label{eq:3}
\end{equation}
The idea of masking channels and selecting only part of them to choose one operation from O (optimizing the $\alpha$ architecture parameter) is shown in Fig.~\ref{fig:alphas_beta}, keeping the symbols consistent with the discussed formulas.
Unfortunately, this modification increases instability in edge selection between nodes.
To counteract this, PC-DARTS introduces edge normalization with an additional parameter $\beta$ to stabilize the network architecture, formulated as:
\begin{align}
	\Psi_{i,j} &= \frac{\exp (\beta_{i,j})}{\Sigma_{i'<j} \exp(\beta_{i',j})} \label{eq:4} \\
	x^{PC}_j &= \sum_{i<j}{(\Psi_{i,j} \cdot f_{i,j}^{PC}(x_i;\mathcal{M}_{i,j}))} \label{eq:5}
\end{align}

\begin{figure}[ht]
	\centering
	\begin{center}
		\includegraphics[width=0.98\columnwidth]{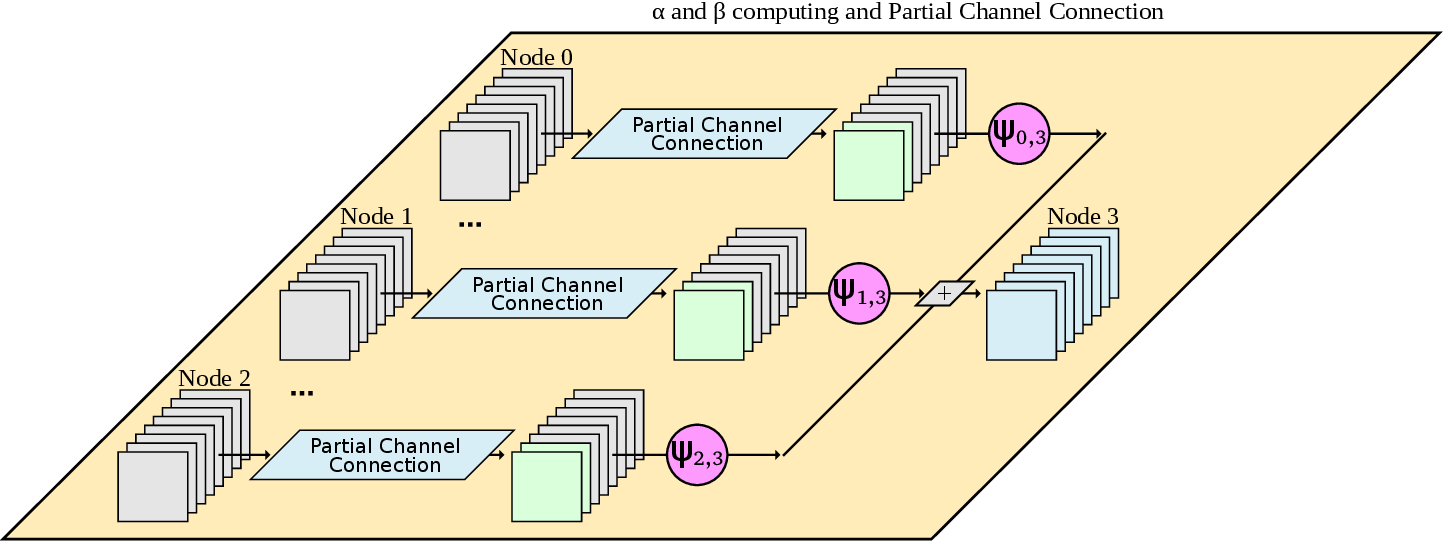} 
	\end{center}
	\caption{A visual representation of the process of searching for node connections that minimize the uncertainty provoked by sampling (refer to Fig.~\ref{fig:alphas_beta}).}
	\label{fig:pc_darts}
\end{figure}

A visual representation of this process is shown in Fig.~\ref{fig:pc_darts}, which follows the previously described notation. Specifically, it highlights the Partial Channel Block and its role in optimizing the $\beta$-cell architecture parameters, as further detailed in Fig.~\ref{fig:alphas_beta}. 

\textbf{We propose an enhancement} to (\ref{eq:4}) by replacing $\beta_{i,j}$ with $\tan(\beta_{i,j})$ and rescaling it, which enhances sparsity by amplifying differences between probabilities, forcing clearer selections between operations. Let's define:
\begin{equation}
\hat{\Psi_{i,j}} = \frac{\exp(\tan(\beta_{i,j})-\max(\tan(\beta_{i,j})))}{\sum_{i'<j} \exp(\tan(\beta_{i',j})-\max(\tan(\beta_{i,j})))}
\label{eq:5.1}
\end{equation}
then, if the highest probability does not exceed 0.5:
\begin{equation}
	\Psi_{i,j} = \hat{\Psi_{i,j}}, ~~ \text{if } \max(\hat{\Psi_{i,j}}) \leq 0.5 \\
	\label{eq:5.2}
	\end{equation}
but, if the highest probability exceeds 0.5 ($\max(\hat{\Psi_{i,j}}) > 0.5$), all values are scaled proportionally to retain a sum of 1 in the following way:
\begin{equation}
	\Psi_{i^*,j} = 0.5, ~ \text{for } i^{*}=\text{argmax}(\hat{\Psi_{i,j}})
	\label{eq:5.2}
\end{equation}
Next, the sum of the rest values (omitting max value) is calculated:
\begin{equation}
	S_{rest} = \sum_{i \neq i^{*}} {\hat{\Psi_{i,j}}}
	\label{eq:5.2}
\end{equation}
and finally (if $\max(\hat{\Psi_{i,j}}) > 0.5$):
\begin{equation}
	\Psi_{i,j} =
	\begin{cases}
		0.5 \frac{\hat{\Psi_{i,j}}}{S_{rest}}, & i \neq i^{*}\\
		0.5, & \text{if } i=i^{*}
	\end{cases}
	\label{eq:5.2}
\end{equation}

This adjustment ensures activation of at most two edges per node, preventing single-edge dominance while preserving differentiability.


Architecture search follows a bi-level optimization process:

\begin{gather}
\underset{\alpha,\beta}{\text{min}} \phantom{x} {\mathcal{L}_{val}(\omega^*(\alpha,\beta,\omega_{U}), \alpha, \beta, \omega_U)} \label{eq:6} \\
\text{s.t.} \phantom{x} \omega^*(\alpha,\beta,\omega_U) = argmin_{\omega} {\mathcal{L}_{train}(\omega, \alpha, \beta, \omega_U)} \label{eq:7}
\end{gather}
where U-Net weights $\omega_U$ remain frozen. We use a one-step approximation to update parameters efficiently:

\begin{equation}
	\omega \leftarrow \omega - \eta_{\omega} \frac{\partial\mathcal{L}_{train}(\omega, \alpha, \beta, \omega_U)}{\partial\omega}
	\label{eq:7.1}
\end{equation}
where in the first step we update the cell weights, and then on the updated cells we update the architecture parameters:
\begin{equation}
	(\alpha, \beta) \leftarrow (\alpha, \beta) - \eta_{\alpha,\beta}\frac{\partial\mathcal{L}_{val}(\omega, \alpha, \beta, \omega_U)}{\partial(\alpha,\beta)}
	\label{eq:7.2}
\end{equation}

To prevent overfitting, we partition training data into \texttt{train\_search\_dt} and \texttt{val\_search\_dt}. 
Architecture parameters ($\alpha, \beta$) are updated on validation data, while cell weights ($\omega$) are trained separately on \texttt{train\_search\_dt}. 
We use \emph{Dice loss} as the optimization objective.


\subsubsection{Discretization}\label{sec:discretization}
After optimization, two architectural parameter sets require discretization: operation selection ($\alpha$) and edge selection ($\beta$). The $\alpha$-discretization phase selects the highest-weighted operation between nodes $(i, j)$, though other operations may still influence the output at node $j$. Discarding non-zero operations can lead to a discretization gap.
Similarly, the $\beta$-discretization phase selects two input edges per node, potentially rejecting edges with small but non-zero probabilities. This issue worsens with increasing cell complexity. While not addressed in this study, acknowledging this limitation is crucial for a comprehensive understanding of the discretization process.\\

\section{Experiments}\label{sec:experiments}

\subsection{Dataset}\label{sec:dataset}

\begin{figure}[th]
	\begin{center}
		\includegraphics[width=0.75\columnwidth]{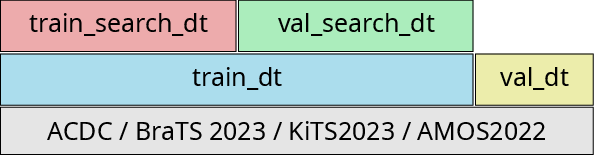} 
	\end{center}
	\caption{The base U-Net architectures were trained utilizing \texttt{train\_dt} and \texttt{val\_dt} sets, effectively serving as reference points in our studies. These datasets were also used in the training of the networks to which the IAC cell had been integrated. Two equally numbered sets, \texttt{train\_search\_dt} and \texttt{val\_search\_dt}, exclusively derived from the \texttt{train\_dt} set, were used in uncovering the adaptive cell’s architecture.}
	\label{fig:dataset}
\end{figure}

Based on recent findings by the authors of \cite{bib:nnUNet2024}, we selected the following datasets for our study:
\begin{enumerate}
	\item ACDC -- heart MRI scans\cite{bib:acdc}, 
	\item BraTS2023 -- brain MRI scans \cite{bib:BraTS-2015,bib:BraTS-Dodatkowy1,bib:BraTS-2023}, 
	\item KiTS2023 -- kidney and renal tumor CT scans\cite{bib:kits21}, 
	\item AMOS2022 -- multi-organ CT \& MRI scans  \cite{bib:amos}.
\end{enumerate}
Networks were trained on multi-channel inputs with corresponding labels. To optimize computation, images were cropped to $128\times128$, reducing background pixels.
Each mask represents binary maps of all relevant classes for each dataset.
To ensure a fair evaluation of our modified U-Net, we avoided dataset augmentation and eliminated any randomness during data preprocessing.
The training dataset was divided into two sections as shown in Fig.~\ref{fig:dataset}.
It consisted of a training set (\texttt{train\_dt}), covering 80\% of the entire dataset, and a validation set (\texttt{val\_dt}) from the remaining 20\%, on which different U-Net models were trained and validated (see Section~\ref{sec:stage_1}).
Furthermore, equal subsets (\texttt{train\_search\_dt} and \texttt{val\_search\_dt}) were extracted from \texttt{train\_dt} and used to search for the optimal cell architecture (see Section~\ref{sec:stage_2}).
After integrating the generated adaptive cell with the pre-trained U-Net, the network was re-trained on \texttt{train\_dt} and validated on \texttt{val\_dt} (see Section~\ref{sec:stage_3}).

\subsection{Adaptive Cell search conditions and implementation details}\label{sec:adaptive_impl}

This section details the experiments conducted to evaluate the efficiency, scalability, and applicability of IAC across multiple U-shaped networks. 
The process consists of three main stages: (I) preparing base networks, (II) searching for IACs, and (III) training the IACs within the network.


\subsubsection{Stage I - Preparing benchmark U-Nets}\label{sec:stage_1}

Our supernet $\boldsymbol{\mathcal{S}}$ follows the U-Net structure shown in Fig.\ref{fig:unet_general}. 
Input images and output labels have a resolution of $128\times128$, with each output channel representing a dataset-specific class (see Section \ref{sec:dataset}). 
This cell-free network, using standard concatenation in skip connections, is trained for 200 epochs on \texttt{train\_dt}. 
Performance is evaluated using the Dice metric on \texttt{val\_dt}, with the best checkpoint selected as the reference model.

Results in Table~\ref{tab:backbone_results} provide a baseline for comparison. 
All networks were trained with the Adam optimizer, a constant learning rate of 0.001, and a batch size of 128. 
This configuration was consistent across all backbones.

Stage I serves as a benchmark to test the generated cells' functionality under different conditions, evaluating various U-shaped architectures and datasets before integrating the adaptive cell.

\begin{table}[h]
	\centering
	\definecolor{headercolor}{gray}{0.9}
	\renewcommand{\arraystretch}{1.1}
	\footnotesize
    \setlength{\tabcolsep}{2pt}
	\caption{Performance of U-Net variants with different backbones using the Dice metric. TR: training set, VAL: validation set.}
	\label{tab:backbone_results}
	\begin{tabular}{c|cc|cc|cc|cc}
		\hline
		\multirow[c]{2}{*}{\begin{tabular}{@{}l@{}} \makebox[0.23\linewidth][r]{DATASET} \\ \hline \makebox[0.23\linewidth][l]{BACKBONE} \end{tabular}} & \multicolumn{2}{c|}{ACDC} & \multicolumn{2}{c|}{AMOS} & \multicolumn{2}{c|}{BRATS} & \multicolumn{2}{c}{KITS}                                 \\
		                                                                                                                                                & TR                     & VAL                       & TR                      & VAL                      & TR & VAL   & TR & VAL   \\ \hline
		Base                                                                                                                                            & 0.965                     & 0.727                     & 0.876                      & 0.595                    & 0.919 & 0.403 & 0.848 & 0.445 \\
		VGG16                                                                                                                                           & 0.96                      & 0.757                     & 0.922                      & 0.626                    & 0.936 & 0.581 & 0.857 & 0.452 \\
		ResNet50                                                                                                                                        & 0.956                     & 0.684                     & 0.904                      & 0.633                    & 0.92  & 0.542 & 0.854 & 0.598 \\
		MobileNetV3-L                                                                                                                               & 0.960                     & 0.722                     & 0.885                      & 0.453                    & 0.273 & 0.254 & 0.841 & 0.072 \\
		EfficientNetV2-S                                                                                                                                & 0.956                     & 0.719                     & 0.926                      & 0.641                    & 0.92  & 0.634 & 0.854 & 0.451 \\
		EfficientNetV2-M                                                                                                                                & 0.966                     & 0.67                      & 0.924                      & 0.707                    & 0.925 & 0.618 & 0.855 & 0.609 \\
		ConvNeXt-Tine                                                                                                                                   & 0.935                     & 0.071                     & 0.917                      & 0.294                    & 0.866 & 0.068 & 0.836 & 0.266 \\
		ConvNeXt-Base                                                                                                                                   & 0.929                     & 0.070                     & 0.909                      & 0.331                    & 0.879 & 0.069 & 0.826 & 0.278 \\

		\hline
	\end{tabular}
\end{table}


\subsubsection{Stage II - Searching Adaptive Cell for appropriate U-Net architecture}\label{sec:stage_2}

Stage II focuses on discovering the optimal Implantable Adaptive Cell (IAC) for enhancing U-Net architectures in medical segmentation. 
A supernet $\mathcal{S}$ (Fig.~\ref{fig:unet_general}) is created, incorporating a continuous representation of candidate cells. 
Pre-trained U-Net weights $\omega_U$ are loaded, while architecture parameters $\alpha$ and $\beta$ are initialized to ones, and cell weights $\omega$ are randomly initialized. 
To ensure compatibility with the trained U-Net, all weights except $\alpha$, $\beta$, and $\omega$ are frozen.
Implementation specifics, correlating to the figures included in our paper, are illustrated in Algorithm~\ref{alg:1} and Algorithm~\ref{alg:2}.
\begin{algorithm}[ht]
	\caption{U-Net adaptive cell search (Stage II).}\label{alg:1}
	\begin{algorithmic}[1]
		\REQUIRE Adaptive cell architecture parameters $\boldsymbol\alpha$ and $\boldsymbol\beta$; IAC weights $\boldsymbol\omega$; Trained vanilla U-Net weights $\boldsymbol\omega_U$, Number of search epochs $\boldsymbol{\mathcal{E}}$; Training dataset \texttt{train\_search\_dt}; Validation dataset \texttt{val\_search\_dt}; Regularization coefficient adjustment schema $\boldsymbol{\lambda_e, e \in \{1, 2, . . . , \mathcal{E}\}}$; Number of steps in one epoch $\boldsymbol{\mathcal{Z} = |train\_search\_dt| = |val\_search\_dt|}$; Loss function $\boldsymbol{\mathcal{L}}$;
		\STATE Construct a supernet $\boldsymbol{\mathcal{S}}$ (see Fig.~\ref{fig:unet_general}) and load U-Net weights $\boldsymbol\omega_U$
		\STATE Initialize architecture parameters $\boldsymbol\alpha$, $\boldsymbol\beta$ with ones and random initialize cells' weights $\boldsymbol\omega$
		\STATE Freeze all weights except $\boldsymbol\alpha$, $\boldsymbol\beta$, $\boldsymbol\omega$ parameters
		\FOR{\textbf{each} $e \in [1, \mathcal{E}]$}
		\FOR{\textbf{each} $z \in [1, \mathcal{Z}]$}		
		\STATE Compute loss value $\boldsymbol{\mathcal{L}_{val\_search\_dt}}$  on the \texttt{val\_search\_dt} based on step forward through the supernet $\mathcal{S}$ (see Algorithm~\ref{alg:2} for more cell's step forward details)
		\STATE \fcolorbox{zolty}{zolty}{\parbox{0.92\linewidth}{Update $\boldsymbol\alpha$ and $\boldsymbol\beta$ by descending $\boldsymbol{\nabla_{\alpha,\beta}\mathcal{L}_{val\_search\_dt}(\alpha,\beta,\omega,\omega_U)}$}}
		\STATE Compute loss value $\boldsymbol{\mathcal{L}_{train\_search\_dt}}$ on the \texttt{train\_search\_dt} based on the updated supernet $\mathcal{S}$ architecture
		\STATE Update $\boldsymbol{\omega}$ by descending $\boldsymbol{\nabla_{\omega}\mathcal{L}_{train\_search\_dt}(\alpha,\beta,\omega,\omega_U)}$
		\ENDFOR
		\ENDFOR
		\STATE \fcolorbox{rozowy}{rozowy}{\parbox{0.92\linewidth}{Derive the IAC architecture based on the learned $\boldsymbol\alpha$,$\boldsymbol\beta$}}
	\end{algorithmic}
\end{algorithm}

A 15-epoch \textit{warm-up} phase is introduced to stabilize searching by pre-optimizing cell weights. 
The actual search phase follows, iterating over \texttt{train\_search\_dt} and \texttt{val\_search\_dt}. 
The loss function $\boldsymbol{\mathcal{L}_{val\_search\_dt}}$ is computed for each batch, updating $\alpha$ and $\beta$ according to Algorithm~\ref{alg:2}. Subsequently, $\boldsymbol{\mathcal{L}_{train\_search\_dt}}$  is used to refine the cell weights $\omega$, as outlined in Algorithm~\ref{alg:1}.
Optimization is conducted using SGD for $\omega$ (learning rate 0.01, no weight decay, momentum 0.0) and Adam for $\alpha$ and $\beta$ (learning rate 0.001). 
A cosine power annealing schedule \cite{bib:sharpDARTS} manages learning rate adjustments. 
The \emph{Dice} loss function is employed, consistent with Stage I.
Due to GPU memory constraints, batch size is set to 128, and the search phase spans 200 epochs.

Although the full 200-epoch search is performed, effective cells can often be identified earlier, around 25 or 50 epochs, as observed in Section~\ref{sec:ablation_genotypes}. 
This suggests that reducing search epochs could save computational resources while maintaining strong results. 
The epoch from which the cell was selected depended on the backbone of the model, and the selection was defined based on ablation studies, although it can be defined arbitrarily.
However, due to the low stability of the DNAS search process, as indicated by the authors of other works, it will be best to check several generated genotypes at Stage II. 
Although our studies do not indicate significant instability in the efficiency of the searched cell architectures, the process of identifying the best cell is dependent on the backbone architecture and the data set.

\begin{algorithm}[ht]
	\caption{U-Net Adaptive Cell Search.}\label{alg:2}
	\begin{algorithmic}[1]
		\REQUIRE Adaptive cell architecture parameters $\boldsymbol\alpha$ and $\boldsymbol\beta(\varphi, \Psi)$; Output node $\boldsymbol j$, its features $\boldsymbol x_j$, and features of all nodes with index smaller than $\boldsymbol{j (x_i \text{ where } i < j)}$; Sampling mask $\mathcal{M}$;
		\setlength{\fboxsep}{1pt}
		\STATE \fcolorbox{zolty}{zolty}{\strut \parbox{0.92\linewidth} {Compute $\boldsymbol x_j$:}}
		\STATE \fcolorbox{zolty}{zolty}{\strut \parbox{0.92\linewidth}{\hspace{5pt}$x_j=0$}}
		\STATE \fcolorbox{zolty}{zolty}{\parbox{0.92\linewidth}{\hspace{5pt}\textbf{for each} $i \in [0, j)$ \textbf{do}}}
		\STATE \fcolorbox{zolty}{zolty}{\parbox{0.92\linewidth}{\hspace{10pt}$x_j += $\colorbox{fioletowy}{$\Psi_{i,j}$} * \colorbox{niebieski}{Compute edge $i$, $j$  for $x_i$ input}}}
		\STATE \fcolorbox{zolty}{zolty}{\parbox{0.92\linewidth}{\hspace{5pt}Return $x_j$}}
		\STATE \fcolorbox{niebieski}{niebieski}{\parbox{0.92\linewidth}{Compute edge $i$, $j$ features for $x_i$ input:}}
		\STATE \fcolorbox{niebieski}{niebieski}{\parbox{0.9\linewidth}
			{\hspace{1pt} \fcolorbox{zielony}{zielony}{\parbox{0.92\linewidth}
					{Multiply features $x_i$ by sampling mask $\mathcal{M}_{i,j}$: $\eta$}}}
		}
		\STATE \fcolorbox{niebieski}{niebieski}{\parbox{0.9\linewidth}
			{\hspace{5pt}\fcolorbox{zielony}{zielony}{\parbox{0.92\linewidth}
					{H = 0}}}
		}
		\STATE \fcolorbox{niebieski}{niebieski}{\parbox{0.9\linewidth}
			{\hspace{5pt}\fcolorbox{zielony}{zielony}{\parbox{0.9\linewidth}
				{\textbf{for each} $o \in \mathcal{O}$ \textbf{do} } } }
		}
		\STATE \fcolorbox{niebieski}{niebieski}{\parbox{0.9\linewidth}
			{\hspace{10pt}\fcolorbox{zielony}{zielony}{\parbox{0.88\linewidth}
			{  $H+=$\colorbox{fioletowy}{$\varphi_{i,j}$}$ * o(\eta)$ } } }  
		}
		\STATE \fcolorbox{niebieski}{niebieski}{\parbox{0.9\linewidth}
			{\hspace{10pt}\fcolorbox{zielony}{szary}{\parbox{0.9\linewidth}
			{Multiply features $x_i$ by sampling mask $(1-\mathcal{M}_{i,j}): \eta$}}}
		}
		\STATE \fcolorbox{niebieski}{niebieski}{\parbox{0.92\linewidth} {\hspace{10pt}Return concat($H$, $\eta$)}}

	\end{algorithmic}
\end{algorithm}


\subsubsection{Stage III - Train U-Net with generated Adaptive Cell}\label{sec:stage_3}

In this final stage, the continuous representation of the Implantable Adaptive Cell is converted into its discrete form accoring to Section~\ref{sec:discretization}. The discrete cell architecture $\boldsymbol{\mathcal{G}}$ is then integrated into the U-Net’s skip connections.

After inserting the cell, we initialize its weights $\omega$ randomly and load the pre-trained U-Net weights $\omega_U$. The U-Net weights remain frozen while the adaptive cell is trained for 200 epochs, following the same hyper-parameters as in Stage I. This step is necessary since, during the search stage, only a single optimization step is performed for the cell weights $\omega$ at each iteration (see Algorithm~\ref{alg:1}). Further details of this stage are outlined in Algorithm~\ref{alg:3}.

\begin{algorithm}[ht]
	\caption{Any U-Net with generated Adaptive Cell.}\label{alg:3}
	\begin{algorithmic}[1]
		\REQUIRE Trained U-Net weights $\boldsymbol{\omega_U}$, Number of search epochs $\boldsymbol{\mathcal{E}}$; Training dataset \texttt{train\_dt}; Validation dataset \texttt{val\_dt}; Generated adaptive cell genotype $\boldsymbol{\mathcal{G}}$;
		\STATE Construct a supernet $\mathcal{S}$ (see Fig.~\ref{fig:unet_general}) based on generated cell genotype $\mathcal{G}$
		\STATE Load specific U-Net weights $\omega_U$ and freeze them
		\STATE Random initialize cells' weights $\omega$
		\STATE \textbf{for each} $e \in [1, \mathcal{E}]$ do:
		\STATE \hspace{\algorithmicindent}Supernet $\mathcal{S}$ training steps on \texttt{train\_dt}
		\STATE \hspace{\algorithmicindent}Supernet $\mathcal{S}$ evaluation steps on \texttt{val\_dt}
		\STATE Derive the final U-Net model with a specific IAC
	\end{algorithmic}
\end{algorithm}

\section{Results}\label{sec:results}
\subsection{Comparative Analysis of Method Effectiveness Across Datasets}

This section evaluates our adaptive cell approach across multiple U-Net architectures, following the protocol outlined in Section~\ref{sec:adaptive_impl} and results from Table~\ref{tab:backbone_results}.
We compared our method with: (1) Attention U-Net \cite{bib:AttentionUNet}, (2) RAUNet \cite{bib:RAUnet}, and (3) UNet++ \cite{bib:UNet++}. These methods modify skip connections, either through attention mechanisms or dense connections, impacting segmentation performance as measured by the Dice Similarity Coefficient (DSC).
Experiments were conducted on four medical imaging datasets: ACDC, AMOS, BRATS, and KITS. Table~\ref{tab:improvements} summarizes DSC improvements over baseline U-Nets from Table~\ref{tab:backbone_results}.

Results indicate that our method outperforms others in feature fusion without major network modifications. The highest DSC gain (0.06) was noted, with AMOS and KITS (0.11) performing best. ACDC showed moderate improvement (0.04), while BRATS remained unchanged due to saturation in the baseline model.
Three key observations emerge from Table~\ref{tab:improvements}: (1) Adding trainable parameters to skip connections does not always yield improvements and may disrupt learned representations. (2) UNet++ consistently performs well due to its dense connections, suggesting potential benefits from aggregating skip connections into a single adaptive cell. (3) Attention-based methods, such as RAUNet and Attention U-Net, indicate that integrating softmax-based attention mechanisms into our search space could further enhance adaptability.

These insights highlight our method's robustness and adaptability in optimizing skip connections for pre-trained U-Nets, offering a scalable alternative to full model redesigns.

\begin{table}[h]
	\centering
	\definecolor{headercolor}{gray}{0.9}
	\renewcommand{\arraystretch}{1.1}
	\footnotesize
	\setlength{\tabcolsep}{4pt}
	\caption{DSC improvement over baseline U-Nets (mean ± variance). Green: improvement, red: worse, yellow: no change. Best results bolded.}
	\label{tab:improvements}

	\begin{tabular}{l|c|c|c|c}
		\hline
		DATASET & PROPOSED                                & ATTENTION                      & RESIDUAL                       & NESTED                         \\
		        & (STAGE III)                             & GATE                           & ATT GATE                       & (U-NET++)                      \\
		\hline
		ACDC    & \cellcolor{zielony}\textbf{0.04 (0.13)} & \cellcolor{zielony}0.01 (0.05) & \cellcolor{rozowy}-0.01 (0.06) & \cellcolor{zielony}0.01 (0.02) \\
		AMOS    & \cellcolor{zielony}\textbf{0.11 (0.09)} & \cellcolor{rozowy}-0.04 (0.10) & \cellcolor{zielony}0.07 (0.08) & \cellcolor{zielony}0.03 (0.03) \\
		BRATS   & \cellcolor{zolty}0.00 (0.01)            & \cellcolor{zolty}0.00 (0.00)   & \cellcolor{rozowy}-0.02 (0.05) & \cellcolor{zolty}0.00 (0.00)   \\
		KITS    & \cellcolor{zielony}\textbf{0.11 (0.09)} & \cellcolor{zielony}0.03 (0.11) & \cellcolor{zielony}0.09 (0.09) & \cellcolor{zielony}0.08 (0.08) \\
		AVERAGE & \cellcolor{zielony}\textbf{0.06 (0.10)} & \cellcolor{zolty}0.00 (0.08)   & \cellcolor{zielony}0.03 (0.08) & \cellcolor{zielony}0.03 (0.05) \\
		\hline
	\end{tabular}
\end{table}

\subsection{Computational Efficiency Comparison}\label{sec:efficiency}

We compare our method’s computational efficiency to three leading NAS techniques in medical image segmentation: NAS-Unet \cite{bib:NAS-Unet}, DiNTS \cite{bib:DiNTS}, and HyperSegNAS \cite{bib:HyperSegNAS}. 
Unlike these full NAS-based models, our approach optimizes only skip connections, drastically reducing computational overhead.

The results of the comparison are shown in Table~\ref{tab:efficiency}. 
The search and final training times were taken from other publications and recalculated to approximate the conditions under which we tested our solution. 
It is clear that our method completes the search phase in hours rather than the days typically required by other NAS approaches. 
This efficiency gain stems from reusing pre-trained encoder-decoder weights while refining skip connections, lowering hardware demands and preserving learned representations—critical advantages in medical imaging with limited datasets and resources. 
Additionally, Stage II search time can be further reduced (see Section V-D).

It is important to note that fully automated NAS methods can yield highly efficient architectures by designing entire networks from scratch. Our approach, however, depends on an existing architecture and may not surpass these solutions in absolute performance. Nonetheless, as observed in \cite{bib:nnUNet2024}, the actual time costs of full NAS methods remain notably higher than those of nnU-Net, underscoring the practical benefits of our method’s more focused strategy.

\begin{table}[t]
	\centering
	\definecolor{headercolor}{gray}{0.3}
	\renewcommand{\arraystretch}{1.1}
	\footnotesize
	\setlength{\tabcolsep}{3pt}
	\caption{Search times (hours) for different NAS methods on four medical imaging datasets. N/A: method not applied. The time of our method is the sum of Stage II and Stage III and is the average time for all backbones. Scaling of GPU Performance: RTX 3090 Ti - 100\% (our GPU), Titan X - 26\%, Tesla V100 - 53\%, RTX 2080 Ti, A100 40GB PCIe - 56\%.}
	\label{tab:efficiency}

	\begin{tabular}{l|ccc|c}
		\hline
		DATASET & \multicolumn{3}{c|}{\begin{tabular}{@{}l@{}} \makebox[0.0\linewidth][c]{OTHER NAS METHODS} \\ \makebox[0.0\linewidth][c]{(FULL ARCHITECTURE SEARCH) } \end{tabular}  } & OUR METHOD                                             \\
		        & NAS-UNET                                                                                                                                                               & DiNTS            & HyperSegNas    &                    \\ \hline
		ACDC    & N/A                                                                                                                                                                    & N/A              & N/A            & $\sim 0.5 (0.1)h$  \\
		AMOS    & N/A                                                                                                                                                                    & $\sim 51.0h^{(2)}$   & $\sim 22.0h^{(3)}$ & $\sim 5.0 (1.4)h$  \\
		BRATS   & $\sim 22.0h^{(1)}$                                                                                                                                                         & N/A              & N/A            & $\sim 11.0 (2.5)h$ \\
		KITS    & N/A                                                                                                                                                                    & $\sim 51.0h^{(2)}$ & N/A            & $\sim 4.5 (2.8)h$  \\
		\hline
	\end{tabular}
	\begin{tablenotes}
		\tiny
		\item[1] $^1$ 72h search, 12h total training, NVIDIA Titan X GPU [$84h * 0.26 = ~22.0h$]\\
		\item[1] $^2$ 18h search, 24h total training, 4 * NVIDIA Tesla V100, 1 * NVIDIA Testla V100 [$18h *(0.53*4) + 24h *0.53 = ~51.0h$]\\
		\item[1] $^3$ 12h search, 15h total training, 2 * RTX 2080 Ti, 1 * RTX 2080 Ti [$12h * (0.56 *2) + 15h*0.56 = ~22.0h$]
	\end{tablenotes}
\end{table}

\subsection{Statistical Analysis of Performance Differences}\label{sec:statistical}
Tables \ref{tab:confidence_ds} and \ref{tab:confidence_models} present the average performance (mean Dice score) and corresponding $95\%$ confidence intervals for (1) the reference networks, (2) networks using the proposed method with IAC Stage III, and (3) networks with connections defined as in U-Net++, aggregated across all datasets (Table \ref{tab:confidence_ds}) and all baseline architectures (Table \ref{tab:confidence_models}).

The paired t-test indicates a significant difference between our method and the baselines $(p \approxeq 0.0008)$, compared to U-Net++ which showed a higher p-value $(p \approxeq 0.0016)$.
Similarly, the Wilcoxon signed-rank test for dependent samples confirms these differences, with our method yields $p < 0.001$ and U-Net++ $p \approxeq 0.002$. 
Our method achieves a mean DSC gain of $0.063$, double that of U-Net++ $(0.0315)$.
Finally, the one-sided Wilcoxon signed-rank test further supports the statistical significance of our results $(Z \approxeq 3.51, p < 0.001)$, demonstrating that the median difference is significantly greater than zero.
Focusing on individual datasets, we observe:
\begin{itemize}
	\item ACDC: Our solution shows the highest average improvement but with high variability, reducing its significance, while U-Net++ offers consistent yet modest gains.
	\item AMOS: Our method significantly improves performance, whereas U-Net++ only reaches borderline significance.
	\item BRATS: No improvement (neither IAC nor U-Net++) is statistically significant, likely due to performance saturation in the baseline model.
	\item KITS: Our solution again outperforms the alternatives with higher significance.
\end{itemize}
The large confidence intervals observed in Table \ref{tab:confidence_models} reflect dataset heterogeneity rather than instability of our method. 
These findings align with our observations that performance varies significantly across different architectures and datasets, but the mean improvement remains consistently positive. 
Future work could investigate how dataset-specific characteristics or U-Net architectures impact the identified architectures of IAC using DNAS, as well as optimize search strategies to mitigate variability

\begin{table}[h]
	\centering
	\definecolor{headercolor}{gray}{0.9}
	\renewcommand{\arraystretch}{1.1}
	\footnotesize
	\setlength{\tabcolsep}{4pt}
	\caption{Performance comparison across datasets.  Values represent mean DSC, with 95\% confidence intervals in parentheses. Values exceeding Baseline U-Net are highlighted in green, lower ones in yellow.}
	\label{tab:confidence_ds}

	\begin{tabular}{l|c|c|c}
		\hline
		DATASET & BASELINE & IAC & U-NET++ \\
		\hline
		
		ACDC    & 0.83 (0.64; 1.02) & \cellcolor{zielony} \textbf{0.87 (0.78; 0.96)} & \cellcolor{zielony} 0.84 (0.66; 1.03) \\
		AMOS    & 0.40 (0.20; 0.59) & \cellcolor{zielony} \textbf{0.50 (0.34; 0.67)} & \cellcolor{zielony} 0.42 (0.23; 0.61) \\
		BRATS   & 0.85 (0.84; 0.86) & \cellcolor{zolty}   0.85 (0.83; 0.86) & \cellcolor{zolty}   0.85 (0.84; 0.86) \\
		KITS    & 0.40 (0.24; 0.55) & \cellcolor{zielony} \textbf{0.50 (0.36; 0.65)} & \cellcolor{zielony} 0.48 (0.36; 0.59) \\
		OVERALL & 0.62 (0.51; 0.72) & \cellcolor{zielony} \textbf{0.68 (0.60; 0.76)} & \cellcolor{zielony} 0.65 (0.56; 0.74) \\
		\hline
	\end{tabular}
\end{table}

\begin{table}[h]
	\centering
	\definecolor{headercolor}{gray}{0.9}
	\renewcommand{\arraystretch}{1.1}
	\footnotesize
	\setlength{\tabcolsep}{2pt}
	\caption{Performance comparison across architectures.  Values represent mean DSC, with 95\% confidence intervals in parentheses. Values exceeding Baseline U-Net are highlighted in green, lower ones in yellow.}
	\label{tab:confidence_models}

	\begin{tabular}{l|c|c|c}
		\hline
		MODEL & BASELINE & IAC & U-NET++ \\
		\hline
		BASE &	          0.65(0.23; 1.08)  & \cellcolor{zielony}\textbf{0.74(0.46; 1.01)}  & \cellcolor{zielony}0.67(0.29; 1.06) \\
		VGG16&	          0.71(0.34; 1.07)  & \cellcolor{zielony}\textbf{0.76(0.50; 1.02)}  & \cellcolor{zielony}0.75(0.48; 1.02) \\
		ResNet50&	      0.73(0.43; 1.03)  & \cellcolor{zielony}\textbf{0.76(0.52; 1.00)}  & \cellcolor{zolty} 0.73(0.43; 1.03) \\
		MobileNetV3-L&	  0.36(-0.17; 0.89) & \cellcolor{zielony}\textbf{0.57(0.21; 0.94)}  & \cellcolor{zielony}0.42(-0.02; 0.87) \\
		EfficientNetV2-S& 0.71(0.37; 1.06)  & \cellcolor{zielony}\textbf{0.76(0.52; 1.00)}  & \cellcolor{zielony}0.75(0.49, 1.01) \\
		EfficientNetV2-M& 0.75(0.49; 1.01)  & \cellcolor{zielony}\textbf{0.77(0.53; 1.00)}  & \cellcolor{zielony}0.76(0.51; 1.09) \\
		ConvNeXt-Tiny&	  0.51(-0.13; 1.15) & \cellcolor{zielony}\textbf{0.54(-0.01; 1.08)} & \cellcolor{zielony}\textbf{0.56(-0.04; 1.16)} \\
		ConvNeXt-Base&	  0.51(-0.13; 1.15) & \cellcolor{zielony}\textbf{0.55(0.00; 1.11)}  & \cellcolor{zielony}0.54(-0.08; 1.17) \\
		\hline
	\end{tabular}
\end{table}


\subsection{Ablation study}\label{sec:ablation}

In the following section, we present the results of our ablation studies. Section~\ref{sec:ablation_genotypes} examines the performance of the cells obtained at different stages of the search process, helping to determine if the process truly requires a fixed duration of 200 epochs. In Section~\ref{sec:ablation_network}, we analyze the training curve of the introduced cells, discussing potential overfitting issues and assessing the impact of the baseline U-Net architecture and its learned representations on the training of the adaptive cell.


\subsubsection{The influence of different cell genotypes}\label{sec:ablation_genotypes}

During the search process, different Adaptive Cell genotypes emerge, shaped by the network architecture and initial operation weights. Their effectiveness varies, highlighting the importance of selecting optimal skip connection operations. Table~\ref{tab:ac_results1} presents results for various genotypes integrated and trained per Stage III. Detailed diagrams of the genotypes of all cells within the search process for each U-Net (with various backbones) are available on the project's website https://gitlab.com/emil-benedykciuk/u-net-darts-tensorflow.

Table~\ref{tab:improvements} indicates that all tested methods performed relatively weakly across the ACDC and BRATS datasets, highlighting the need for regularization to stabilize training and improve results \cite{bib:Beta-Decay}. 
Notably, the ACDC dataset exhibited considerable instability during Stage II - characterized by occasional yet significant improvements (see Table~\ref{tab:ac_results1}), which is likely attributable to its limited size and propensity IAC for overfitting. 
As shown in Table~\ref{tab:ac_results1}, strong candidates are frequently identified within 25 steps, suggesting that a shorter search phase may reduce computational costs without sacrificing performance. Although data augmentation was deliberately omitted to maintain fairness, its potential benefits in enhancing our method performance and generalization warrant further investigation.

\begin{table}[htbp!]
	\centering
	\definecolor{headercolor}{gray}{0.8}
	\definecolor{cellcolor}{gray}{1.0}
	\caption{Results of different adaptive cell genotypes. The table shows Dice coefficients for U-Nets with various backbones and adaptive cell genotypes from Stage II (EP) and processed in Stage III. Values exceeding Stage I reference (BSLN) are highlighted in green, lower ones in red, and the best in bold.}
	\label{tab:ac_results1}
	\footnotesize
	\setlength{\tabcolsep}{2pt}
	\renewcommand{\arraystretch}{1.2}
	\begin{tabular}{l|l|c|c|c|c|c|c}
		\hline
		DATA & BACKBONE          & BSLN           & EP 25                             & EP 50                             & EP 100                            & EP 150                            & EP 200                            \\
		\hline
		        & Base              & \textbf{0.919} & \cellcolor{rozowy}0.915           & \cellcolor{rozowy}0.915           & \cellcolor{rozowy}0.912           & \cellcolor{rozowy}0.912           & \cellcolor{rozowy}0.911           \\
		        & VGG16             & \textbf{0.936} & \cellcolor{rozowy}0.931           & \cellcolor{rozowy}0.925           & \cellcolor{rozowy}0.932           & \cellcolor{rozowy}0.934           & \cellcolor{rozowy}0.930           \\
		        & ResNet50          & \textbf{0.921} & \cellcolor{rozowy}0.914           & \cellcolor{rozowy}0.913           & \cellcolor{rozowy}0.914           & \cellcolor{rozowy}0.914           & \cellcolor{rozowy}0.916           \\
		ACDC    & MobileNetV3-L & 0.273          & \cellcolor{zielony}0.375          & \cellcolor{zielony}0.401          & \cellcolor{zielony}0.506          & \cellcolor{zielony}0.375          & \cellcolor{zielony}\textbf{0.619} \\
		        & EfficientNetV2-S  & 0.920          & \cellcolor{rozowy}0.919           & \cellcolor{zielony}\textbf{0.924} & \cellcolor{zolty}0.920            & \cellcolor{zielony}0.923          & \cellcolor{rozowy}0.919           \\
		        & EfficientNetV2-M  & \textbf{0.925} & \cellcolor{rozowy}0.923           & \cellcolor{rozowy}0.924           & \cellcolor{rozowy}0.924           & \cellcolor{rozowy}0.923           & \cellcolor{rozowy}0.922           \\
		        & ConvNeXt-Tiny     & \textbf{0.866} & \cellcolor{rozowy}0.632           & \cellcolor{rozowy}0.397           & \cellcolor{rozowy}0.826           & \cellcolor{rozowy}0.768           & \cellcolor{rozowy}0.751           \\
		        & ConvNeXt-Base     & 0.879          & \cellcolor{rozowy}0.717           & \cellcolor{zielony}\textbf{0.883} & \cellcolor{rozowy}0.676           & \cellcolor{rozowy}0.760           & \cellcolor{rozowy}0.401           \\
		\hline
		        & Base              & 0.403          & \cellcolor{zielony}0.506          & \cellcolor{zielony}0.486          & \cellcolor{zielony}0.528          & \cellcolor{zielony}\textbf{0.549} & \cellcolor{zielony}0.539          \\
		        & VGG16             & 0.581          & \cellcolor{zielony}\textbf{0.632} & \cellcolor{zielony}0.609          & \cellcolor{zielony}0.627          & \cellcolor{zielony}0.605          & \cellcolor{zielony}0.632          \\
		        & ResNet50          & 0.541          & \cellcolor{rozowy}0.530           & \cellcolor{zielony}0.584          & \cellcolor{zielony}0.553          & \cellcolor{zielony}0.558          & \cellcolor{zielony}\textbf{0.634} \\
		AMOS    & MobileNetV3-L & 0.255          & \cellcolor{zielony}0.536          & \cellcolor{zielony}0.474          & \cellcolor{zielony}0.503          & \cellcolor{zielony}\textbf{0.540} & \cellcolor{zielony}0.419          \\
		        & EfficientNetV2-S  & \textbf{0.634} & \cellcolor{rozowy}0.595           & \cellcolor{rozowy}0.598           & \cellcolor{rozowy}0.591           & \cellcolor{rozowy}0.620           & \cellcolor{rozowy}0.622           \\
		        & EfficientNetV2-M  & 0.618          & \cellcolor{zielony}0.644          & \cellcolor{zielony}0.629          & \cellcolor{zielony}0.623          & \cellcolor{zielony}\textbf{0.660} & \cellcolor{zielony}0.633          \\
		        & ConvNeXt-Tiny     & 0.071          & \cellcolor{zielony}0.175          & \cellcolor{zielony}0.162          & \cellcolor{zielony}0.176          & \cellcolor{zielony}\textbf{0.196} & \cellcolor{zielony}\textbf{0.196} \\
		        & ConvNeXt-Base     & 0.070          & \cellcolor{zielony}0.164          & \cellcolor{zielony}0.183          & \cellcolor{zielony}0.163          & \cellcolor{zielony}0.176          & \cellcolor{zielony}\textbf{0.185} \\
		\hline
		        & Base              & 0.848          & \cellcolor{zielony}0.850          & \cellcolor{zielony}0.850          & \cellcolor{zielony}0.850          & \cellcolor{zielony}\textbf{0.852} & \cellcolor{zielony}0.852          \\
		        & VGG16             & \textbf{0.857} & \cellcolor{rozowy}0.855           & \cellcolor{rozowy}0.856           & \cellcolor{rozowy}0.855           & \cellcolor{rozowy}0.855           & \cellcolor{rozowy}0.852           \\
		        & ResNet50          & 0.855          & \cellcolor{zielony}0.855          & \cellcolor{zielony}0.855          & \cellcolor{zielony}\textbf{0.858} & \cellcolor{rozowy}0.854           & \cellcolor{zolty}0.855            \\
		BRATS   & MobileNetV3-L & 0.841          & \cellcolor{zielony}0.842          & \cellcolor{zielony}0.848          & \cellcolor{zielony}0.850          & \cellcolor{zielony}\textbf{0.851} & \cellcolor{zielony}0.844          \\
		        & EfficientNetV2-S  & 0.854          & \cellcolor{zielony}0.857          & \cellcolor{zielony}0.856          & \cellcolor{zielony}\textbf{0.858} & \cellcolor{zielony}0.855          & \cellcolor{zielony}0.855          \\
		        & EfficientNetV2-M  & 0.855          & \cellcolor{zielony}0.856          & \cellcolor{zielony}0.856          & \cellcolor{zielony}0.856          & \cellcolor{zielony}\textbf{0.857} & \cellcolor{zielony}0.856          \\
		        & ConvNeXt-Tiny     & \textbf{0.836} & \cellcolor{rozowy}0.834           & \cellcolor{rozowy}0.835           & \cellcolor{rozowy}0.820           & \cellcolor{rozowy}0.808           & \cellcolor{rozowy}0.834           \\
		        & ConvNeXt-Base     & \textbf{0.826} & \cellcolor{rozowy}0.794           & \cellcolor{rozowy}0.758           & \cellcolor{rozowy}0.8133          & \cellcolor{rozowy} 0.785          & \cellcolor{rozowy} 0.810          \\
		\hline
		        & Base              & 0.445          & \cellcolor{zielony}0.606          & \cellcolor{zielony}0.467          & \cellcolor{zielony}\textbf{0.630} & \cellcolor{zielony}0.625          & \cellcolor{zielony}0.595          \\
		        & VGG16             & 0.452          & \cellcolor{zielony}\textbf{0.603} & \cellcolor{zielony}0.456          & \cellcolor{rozowy}0.429           & \cellcolor{zielony}0.602          & \cellcolor{rozowy}0.447           \\
		        & ResNet50          & 0.598          & \cellcolor{zielony}\textbf{0.631} & \cellcolor{zielony}0.612          & \cellcolor{zielony}0.622          & \cellcolor{zielony}0.608          & \cellcolor{zielony}0.601          \\
		KITS    & MobileNetV3-L & 0.071          & \cellcolor{zielony}0.276          & \cellcolor{zielony}0.288          & \cellcolor{zielony}0.280          & \cellcolor{zielony}\textbf{0.290} & \cellcolor{zielony}0.091          \\
		        & EfficientNetV2-S  & 0.451          & \cellcolor{zielony}0.616          & \cellcolor{zielony}0.607          & \cellcolor{zielony}\textbf{0.640} & \cellcolor{zielony}0.601          & \cellcolor{zielony}0.611          \\
		        & EfficientNetV2-M  & 0.609          & \cellcolor{zielony}\textbf{0.625} & \cellcolor{zielony}0.618          & \cellcolor{zielony}0.618          & \cellcolor{zielony}0.615          & \cellcolor{zielony}0.618          \\
		        & ConvNeXt-Tiny     & 0.266          & \cellcolor{rozowy}0.256           & \cellcolor{rozowy}0.262           & \cellcolor{rozowy}0.263           & \cellcolor{rozowy}0.235           & \cellcolor{zielony}\textbf{0.287} \\
		        & ConvNeXt-Base     & 0.278          & \cellcolor{zielony}0.283          & \cellcolor{zielony}0.312          & \cellcolor{zielony}0.285          & \cellcolor{zielony}\textbf{0.322} & \cellcolor{zielony}0.291          \\
		\hline
	\end{tabular}
\end{table}

Our method improved network performance in $72\%$ of cases ($83\%$ excluding ACDC), outperforming Attention Gate ($47\%$, $50\%$ without ACDC) and RA-UNet ($59\%$, $75\%$ without ACDC), though slightly behind UNet++ ($84\%$, $88\%$ without ACDC). 
UNet++'s consequence stems from enhanced feature transfer in skip connections, making gradient flow more stable.
However, as Table~\ref{tab:improvements} shows, our method yields greater overall performance improvements. Additionally, our solution showed the largest improvement over the baseline network in 17 out of 32 ($~53\%$) test cases, where U-Net++ achieved it for 11 out of 32 test cases ($~34\%$).


\subsubsection{Analysis of learning curves across different IAC genotypes}\label{sec:ablation_network}

\begin{figure*}[!ht]
	\centering
	\begin{tabular}{cccc}
		\includegraphics[width=0.23\textwidth]{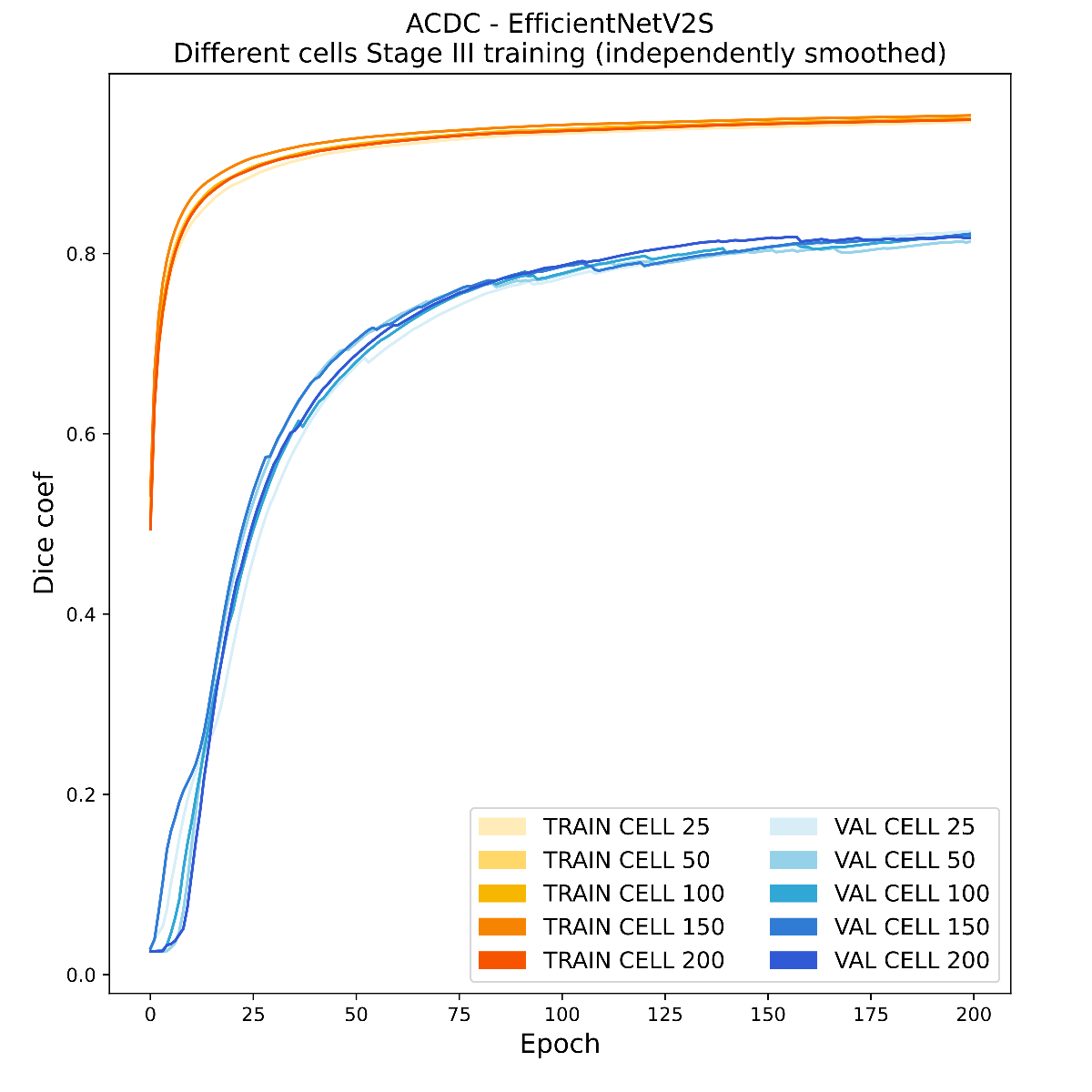} 
		\includegraphics[width=0.23\textwidth]{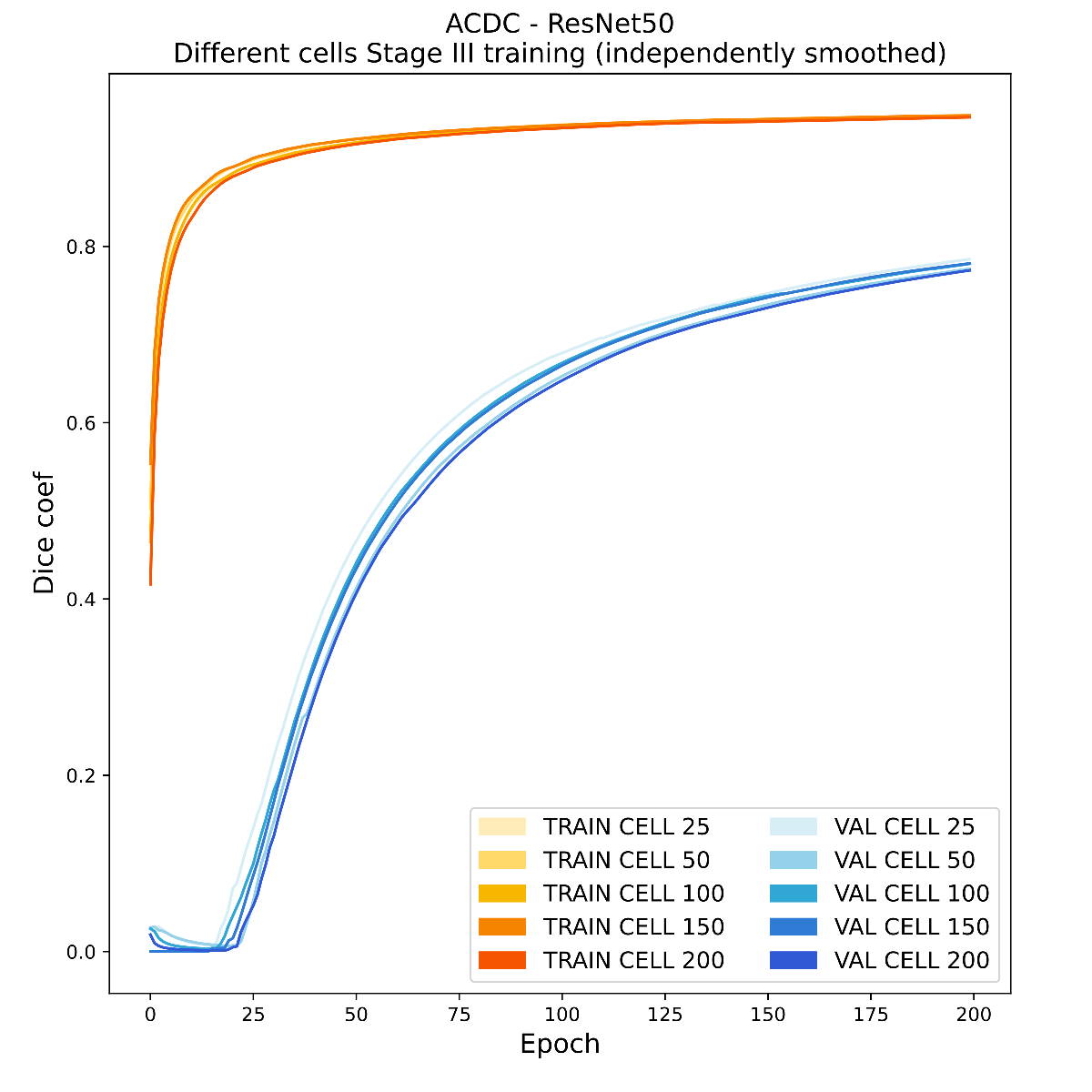} 
		\includegraphics[width=0.23\textwidth]{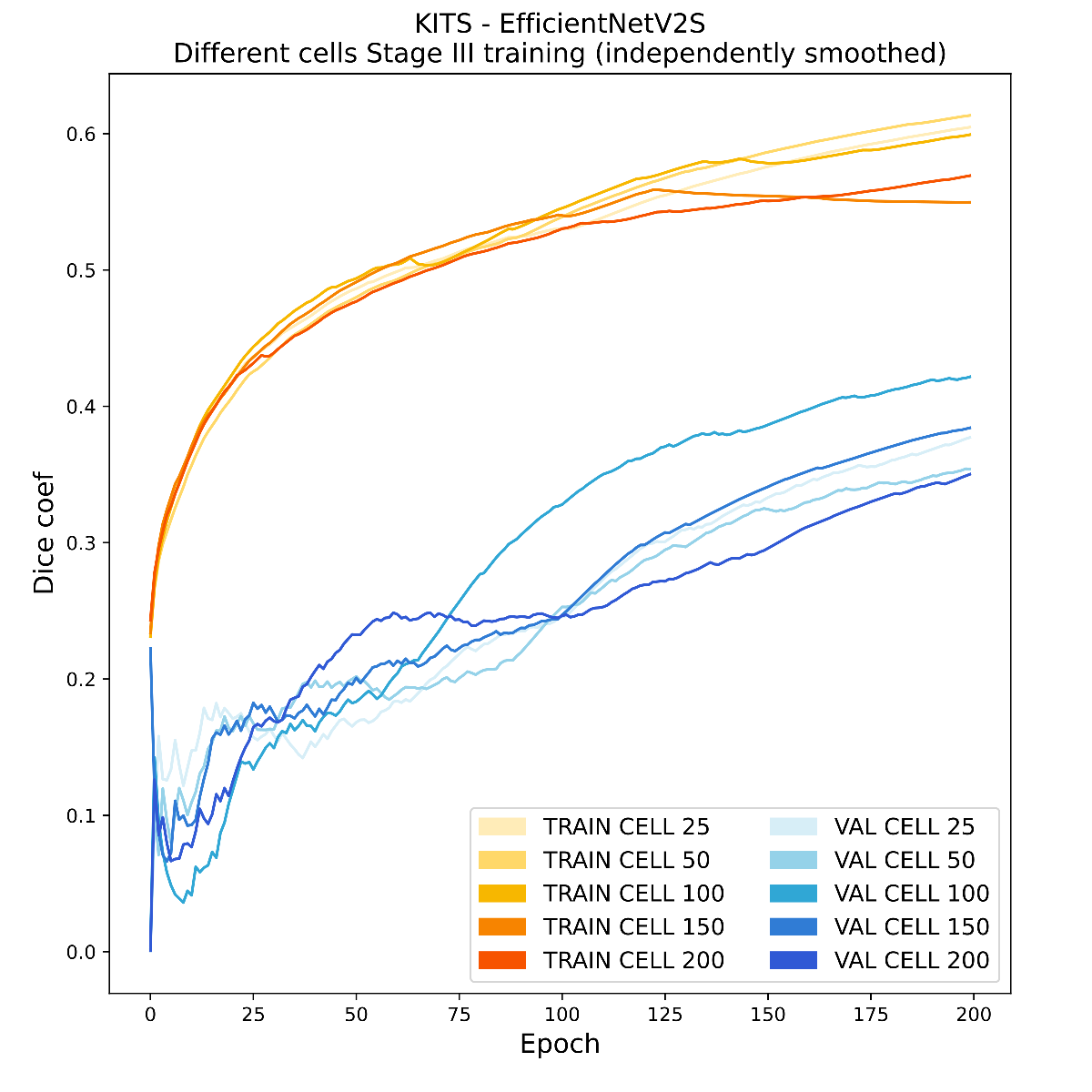} 
		\includegraphics[width=0.23\textwidth]{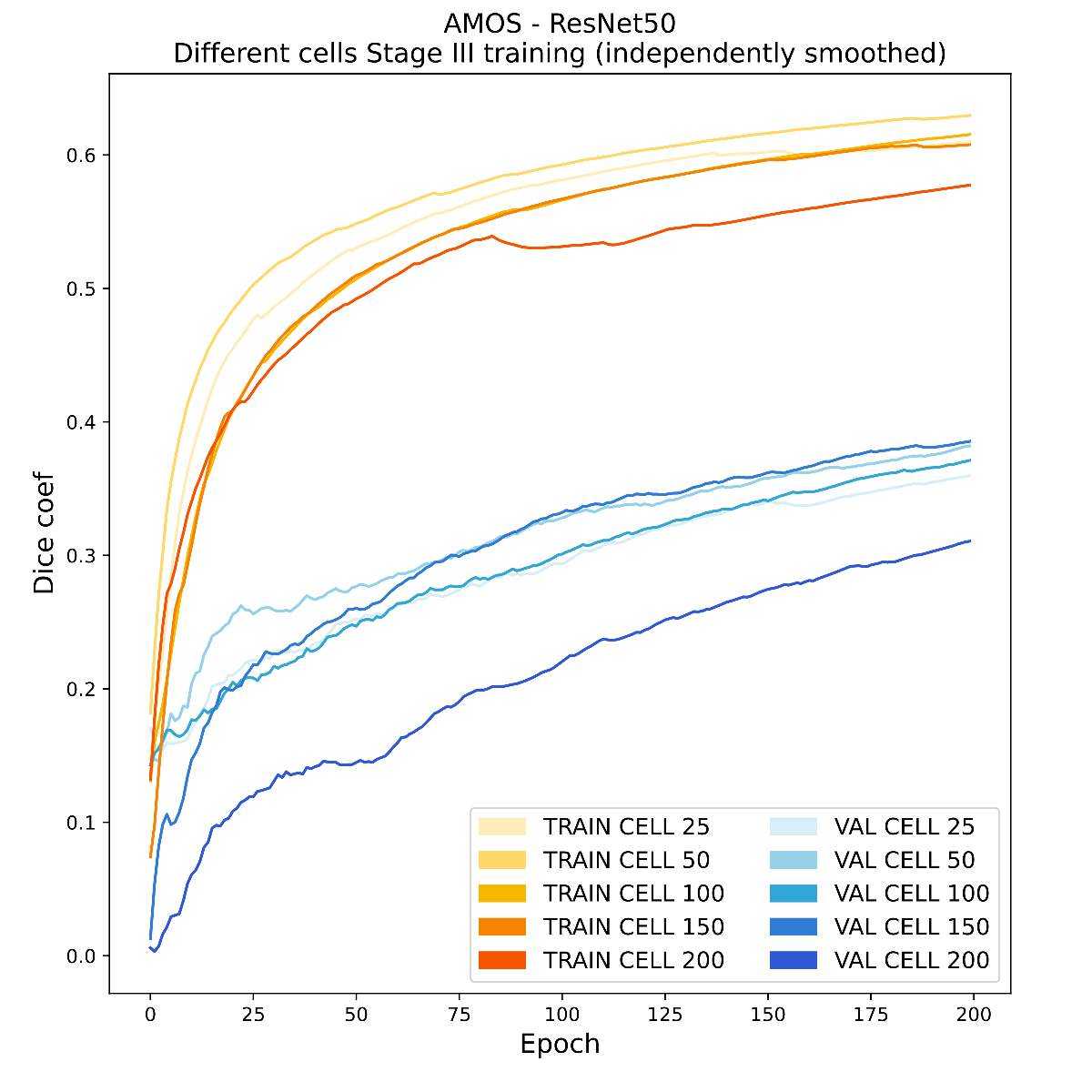} 
	\end{tabular}
	\caption{Example learning curve graphs for selected architectures with discretized cells obtained during training in Stage III. For each architecture, 5 cell genotypes were selected from epochs in Stage II. In each graph, curves have been drawn for the \texttt{train\_dt} (TRAIN CELL, in red–yellow color) and \texttt{val\_dt} (VAL CELL, in bluish color) datasets.}
	\label{fig:curves}
\end{figure*}

Learning curves across different IAC genotypes exhibit similar trends within the same backbone and dataset, despite variations in loss values. This suggests that while adaptive cells enhance performance, their impact on learning behavior is secondary to the pre-trained network architecture.
As shown in Fig.~\ref{fig:curves}, learning dynamics are primarily shaped by the encoder’s feature extraction capability, with skip connections contributing incrementally. Despite structural differences, most genotypes perform similarly, indicating that search-stage variations often refine rather than radically alter feature fusion.

\section{Conclusions}\label{sec:conclusions}

This study shows that optimizing skip connections with Implantable Adaptive Cells (IACs) enhances segmentation performance while reducing computational cost. 
Unlike traditional Neural Architecture Search methods that require training entirely new architectures from scratch,
our method refines pre-trained U-Nets by injecting adaptive modules into their skip connections.
This architecture refinement strategy allows for efficient model adaptation, leveraging pre-existing learned representations rather than discarding them.

Experimental results on medical imaging datasets (ACDC, AMOS, BraTS, and KiTS) confirm that IACs consistently outperform other skip connection configurations, including U-Net++, Attention U-Net, and RA-UNet, with statistically significant DSC gains of 6.3pp on average, nearly twice that of U-Net++.
Paired t-tests and Wilcoxon tests confirm that improvements are non-random.

\textbf{Clinical Implications.}
Accurate segmentation plays a crucial role in clinical workflows, impacting tumor delineation, organ boundary detection, and lesion quantification. By enhancing segmentation accuracy while maintaining computational efficiency, our approach presents a practical alternative to full model searching of training, making it suitable for real-time medical imaging applications. Potential clinical applications of IAC-enhanced U-Nets include: (1) MRI-guided interventions, where improved segmentation accuracy can aid in surgical navigation, (2) Radiotherapy planning, where precise delineation of tumors is essential for dose optimization, (3) Automated radiology workflows, integrating AI-based segmentation into CAD systems to support clinical decision-making.
Furthermore, hospitals and research centers often lack the computational power and large annotated datasets required for deep learning model retraining. By enabling model refinement without requiring full retraining, IACs could help democratize access to advanced AI-driven segmentation tools, making them more accessible in real-world clinical environments.

\textbf{Limitations and Future Work.}
While our study demonstrates the effectiveness of IAC-enhanced U-Nets, several challenges remain. The experiments were conducted on publicly available datasets, and future work should focus on prospective clinical validation, multi-center datasets to evaluate robustness across different imaging protocols. 
Future work should extend IACs to 3D segmentation and Transformer-based models like Swin-UNet \cite{bib:swin-unet} or UNETR \cite{bib:unetr}.

DNAS search stability remains a challenge due to sensitivity to architectural choices. 
Regularization techniques, like Bayesian optimization or parameter pruning, could enhance efficiency.
Additionally, integrating attention mechanisms may improve feature selection for fine-grained structures, further refining IACs for broader medical imaging applications.

\bibliographystyle{IEEEtran}
\bibliography{references}

\end{document}